\documentclass[10pt,twocolumn,letterpaper]{article}

\usepackage{iccv}
\usepackage{times}
\usepackage{epsfig}
\usepackage{graphicx}
\usepackage{amsmath}
\usepackage{amssymb}
\usepackage{xcolor}
\usepackage{threeparttable}
\usepackage{pifont}
\newcommand{\cmark}{\ding{51}}%
\newcommand{\xmark}{\ding{55}}%
\usepackage{booktabs}

\usepackage{mathtools}

\usepackage{listings}
\definecolor{codegreen}{rgb}{0,0.5,0}
\definecolor{codeblue}{rgb}{0.25,0.5,0.5}
\definecolor{codegray}{rgb}{0.6,0.6,0.6}
\usepackage{multirow}

\usepackage[pagebackref=true,breaklinks=true,letterpaper=true,colorlinks,bookmarks=false]{hyperref}

\graphicspath{{figure/}}

\iccvfinalcopy 

\def\iccvPaperID{2831} 

\ificcvfinal\fi
\begin{document}

\title{Interlaced Sparse Self-Attention for Semantic Segmentation}
\author{
    Lang Huang$^{1*}$, Yuhui Yuan$^{2,3,4}$\thanks{Equal contribution.},
    Jianyuan Guo$^1$, Chao Zhang$^1$, Xilin Chen$^{3,4}$, Jingdong Wang$^2$\\
    $^1$Key Laboratory of Machine Perception (MOE), Peking University\quad
    $^2$Microsoft Research Asia\\
    $^3$Institute of Computing Technology, CAS\quad
    $^4$University of Chinese Academy of Sciences \\
    {\tt \small \{laynehuang, jyguo, c.zhang\}@pku.edu.cn,
    \{yuhui.yuan, jingdw\}@microsoft.com, xlchen@ict.ac.cn}
}

\maketitle



\begin{abstract}
In this paper,
we present a so-called interlaced sparse self-attention approach
to improve the efficiency 
of the \emph{self-attention} mechanism for semantic segmentation.
The main idea is that we factorize the dense affinity matrix
as the product of two sparse affinity matrices.
There are two successive attention modules 
each estimating a sparse affinity matrix.
The first attention module is used 
to estimate the affinities within 
a subset of positions that 
have long spatial interval distances
and the second attention module is 
used to estimate the affinities within
a subset of positions that
have short spatial interval distances.
These two attention modules are designed
so that each position is able to receive the information 
from all the other positions.
In contrast to the original self-attention module, 
our approach decreases the computation and memory complexity 
substantially especially when processing high-resolution feature maps.
We empirically verify the effectiveness of our approach
on six challenging semantic segmentation benchmarks.
\end{abstract}
\vspace{-0.5cm}
\section{Introduction}
Long-range dependency plays an essential role
for various computer vision tasks.
The conventional deep convolutional neural networks model the
long-range dependencies mainly by stacking multiple convolutions.
According to the recent work~\cite{shen2018factorized},
we need to stack nearly hundreds of consecutive $3\times3$ convolutions
to enable the model capturing the dependencies between any positions
given the input of size $256\times 256$, and we can see that
the \emph{stacking} scheme leads to very deep models that suffers from poor practical value.

\begin{figure}[t]
\includegraphics[width=.475\textwidth]{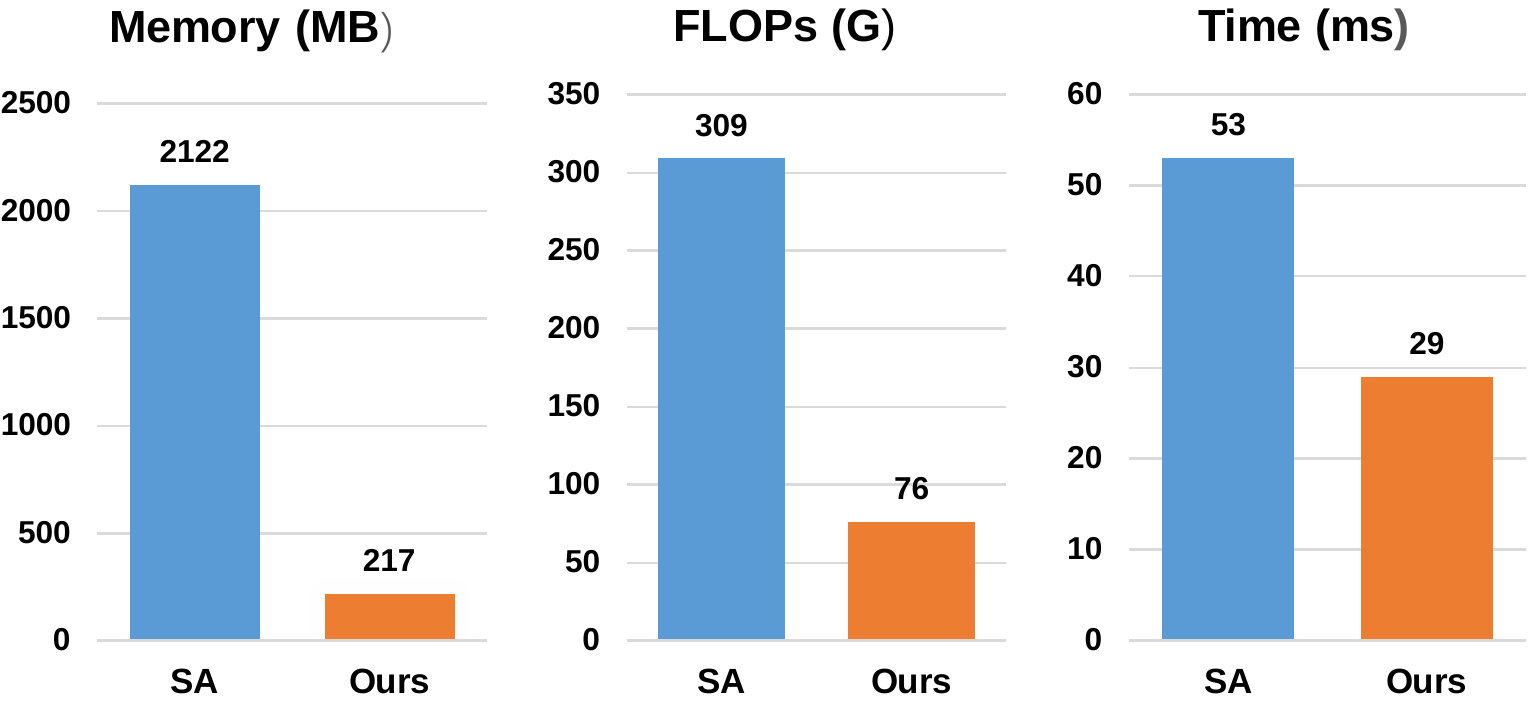}
\caption{\small{GPU memory/FLOPs/Time comparison between self-attention mechanism (SA) and our approach.
All the numbers are tested on a single Titan XP GPU with CUDA$8.0$ and
an input feature map of $1\times 512\times 128\times 128$
during inference stage.
The lower, the better for all metrics.
It can be seen that our approach only uses $10.2\%$ GPU memory and $24.6\%$ FLOPs while being nearly
$2\times$ faster when compared with the self-attention.
}}
\vspace{-0.6cm}
\label{fig:cost}
\end{figure}

The recent self-attention~\cite{vaswani2017attention}
(\emph{or} non-local~\cite{wang2018non}) mechanism 
proposes to model the long-range dependencies through 
computing the context information
of each output position by attending to all the input positions.
We can model the dependencies between any positions in the input
with a single layer equipped with self-attention mechanism.
The self-attention scheme has been used  
for various vision tasks including:
video understanding~\cite{wang2018non}, object detection~\cite{wang2018non}, 
semantic segmentation~\cite{fu2018dual,OCNet,zhang2019co}
and person re-identification~\cite{liao2018video}.
The computation complexity of self-attention mechanism
is about $\mathcal{O}(N^2)$ given input of size $N$, 
and the cost can become very huge
for the task that requires high-resolution input such
as the object detection~\cite{wang2018non} and 
semantic segmentation~\cite{OCNet}.

The high-resolution inputs,
which are essential for high-performance in various fundamental vision tasks,
take heavy computation/memory cost, 
preventing the potential benefit of self-attention mechanism from practical application.
For example,
the recent works~\cite{fu2018dual,OCNet,zhang2019co} apply the self-attention
for semantic segmentation and require more than
$64$ GB of GPU memory to train a model even for a small batch size such as $8$.
In summary, we argue that how to decrease the computation/memory cost
of self-attention mechanism
is of great practical value for various
vision tasks that are sensitive
to the computation and memory cost.

Considering that the heavy computation and memory cost mainly
comes from the $\mathcal{O}(N^2)$ complexity\footnote{The size of the dense affinity matrix is of size $16384\times16384$ for an input feature map of spatial size $128\times 128$.} of computing
the dense affinity matrix $\mathbf{A}$, where each entry indicates
the similarity between one output position and each input position.
We present a simple yet efficient scheme to factorize the computation
of the dense affinity matrix $\mathbf{A}$ 
as the product of two sparse affinity matrices including
$\mathbf{A}^\mathrm{L}$ and $\mathbf{A}^\mathrm{S}$.
In our implementation, 
both $\mathbf{A}^\mathrm{L}$ and $\mathbf{A}^\mathrm{S}$ are
sparse block affinity matrices and their product 
$\mathbf{A}^\mathrm{S}\mathbf{A}^\mathrm{L}$ is a dense affinity matrix.
Computing these two sparse affinity matrices is much
cheaper and saves large amount of cost.
We theoretically show that the computation complexity of 
our approach is much smaller than the
conventional self-attention mechanism.
We illustrate the advantage of our approach 
over the conventional self-attention
in terms of 
the GPU memory cost (measured by MB),
computation cost (mearsured by GFLOPs), 
and inference cost (measured by ms) in Figure~\ref{fig:cost}.

\begin{figure}[t]
\includegraphics[width=.475\textwidth]{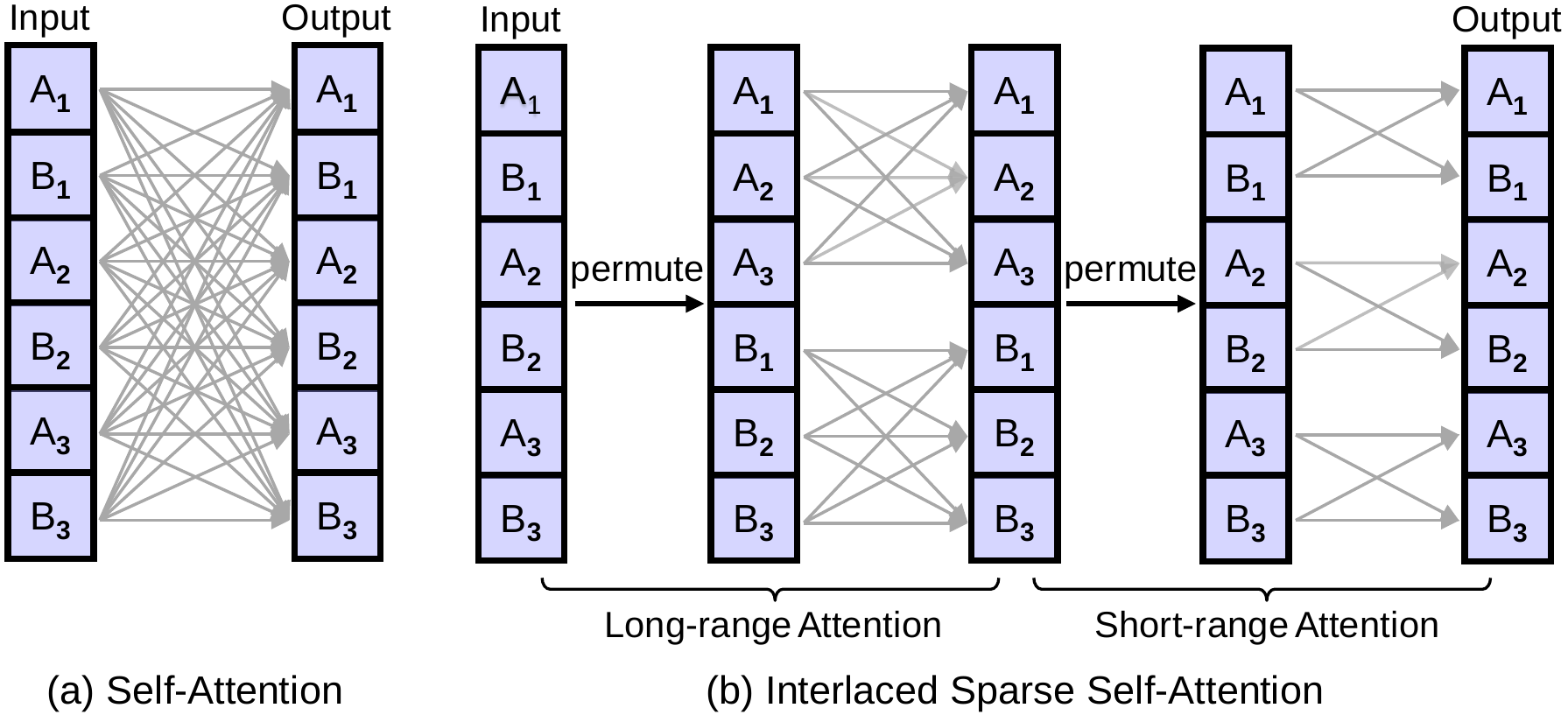}
\caption{\small{
Information propagation path of (a) Self-Attention and (b) Interlaced Sparse Self-Attention.
We use $\mathrm{A}_1,\mathrm{B}_1,...,\mathrm{B}_3$ to represent different positions.
The gray arrows represent the information propagation path from one input position to one output position.
In (a), each output position receives the propagated information
from all the input positions and the connections are fully dense.
In (b), we propagate the information with two steps
and each step only contains very sparse connections.
The first/second permute operation is used to group 
the positions originally have long/short spatial interval distances together.
}
}
\label{fig:connection}
\vspace{-0.5cm}
\end{figure}

The implementation of our approach is 
inspired by the interlacing mechanism~\cite{wiki_interlace}.
First, we divide all the input positions to $\mathcal{Q}$
subsets of equal size, where each set is consisted of 
the $\mathcal{P}$ positions
\footnote{$N=\mathcal{P} \times \mathcal{Q}$, $N$ is the input size.}.
For the long-range attention, we sample 
a position from each subset
to construct a new subset of $\mathcal{Q}$ positions,
and we can construct $\mathcal{P}$ subsets following 
the above sampling strategy.
The positions within each constructed subset are of
long spatial interval distances.
We apply the self-attention on each subset to 
compute the sparse block affinity matrix $\mathbf{A}^\mathrm{L}$.
We propagate information within each sub-group
according to $\mathbf{A}^\mathrm{L}$.
For the short-range attention, we 
directly apply self-attention on the original $\mathcal{Q}$ subsets
to compute the sparse block affinity matrix $\mathbf{A}^\mathrm{S}$.
We then propagate
information within the nearby positions
according to $\mathbf{A}^\mathrm{S}$.
Combining these two attention mechanisms,
we can propagate information from each input position
to all the output positions.
We have illustrated an example for our approach
when processing $1$-D input in Figure~\ref{fig:connection}.


We empirically evaluate the proposed approach
on various vision tasks including 
semantic segmentation, object detection and instance segmentation.
In summary, the contributions of this paper are summarized as below,

\begin{itemize}
    \item We present an \emph{interlaced sparse self-attention}
    scheme to capture the dense long-range dependencies more
    efficiently.
    \item We demonstrate the effectiveness of the \emph{interlaced sparse self-attention}
    scheme on semantic segmentation and achieve similar or
    even better performance compared to the conventional self-attention mechanism.
    \item We empirically compare the \emph{interlaced sparse self-attention} scheme
    with other mechanisms (such as CGNL~\cite{yue2018cgnl}, RCCA~\cite{huang2018ccnet}) 
    to illustrate the advantages of the proposed approach.
\end{itemize}

\section{Related Work}

\paragraph{Self-Attention/Non-local.}
The self-attention~\cite{vaswani2017attention} is originally proposed to solve the machine translation,
and the following work~\cite{wang2018non} further proposed the non-local neural network
for various tasks such as video classification,
object detection and instance segmentation.
\cite{hu2018relation} also applies the self-attention mechanism to model the relations between
the objects for better object detection.
The recent~\cite{fu2018dual,OCNet,zhang2019co} applies the similar mechanism
for semantic segmentation and achieves good performance.
Our work is closely related to the above works
and we are mainly interested in improving the efficiency
of the self-attention when processing the high-resolution input.

\paragraph{CGNL/RCCA.}
The recent works~\cite{yue2018cgnl,huang2018ccnet,A2Net,NIPS2018_7456,li2018beyond} 
all attempt to improve the efficiency of self-attention scheme
and propose various solutions.
For example, the CGNL~\cite{yue2018cgnl} (Compact Generalized Non-local) applies the Taylor series of the RBF kernel function
to approximate the pair-wise similarities,
RCCA~\cite{huang2018ccnet} (Recurrent Criss-Cross Attention) applies two consecutive criss-cross attention to approximate
the original self-attention scheme
and 
A$^2$-Net~\cite{A2Net} applies a set of global representations to propagate the information 
between all the positions
more efficiently.
Compared with the above works,
our work is more simple and easy to implement for various tasks that
originally depend on self-attention mechanism.
We also empirically verify the advantages of our approach over
both CGNL and RCCA.

\begin{figure*}[t]
\includegraphics[width=\textwidth]{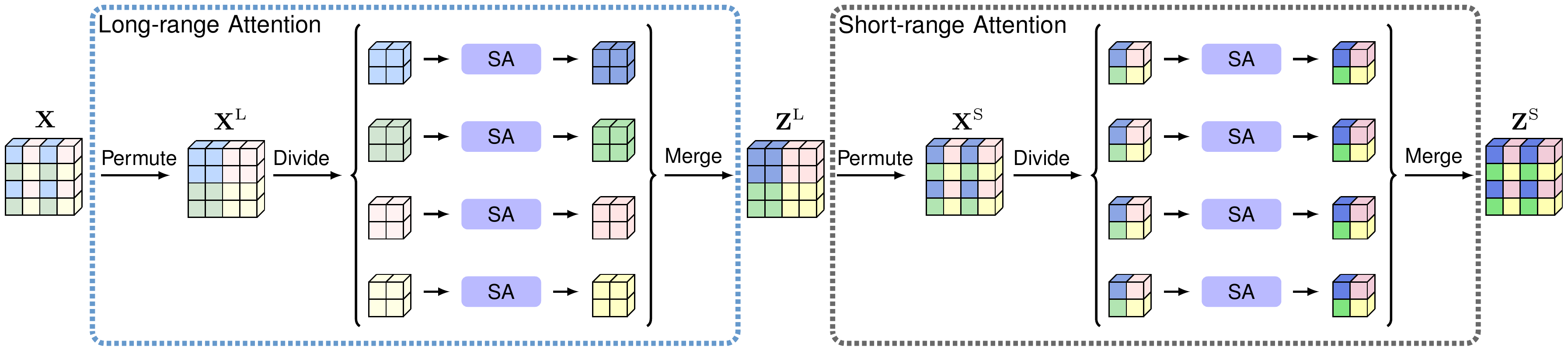}
\caption{\small{
\textbf{Interlaced sparse self-attention}.
Our approach is consisted of a 
long-range attention module and a short-range attention module.
The feature map in the left-most/right-most is the input/output.
First, we colour the input feature map $\mathbf{X}$ with four different colors.
We can see that there are $4$ local groups and each group is
consisted of four different colors.
For the long-range attention, we permute and group (divide) all the positions 
with the same color having long spatial interval distances together in $\mathbf{X}$,
which outputs $\mathbf{X}^{\mathbf{L}}$.
Then, we divide the $\mathbf{X}^{\mathbf{L}}$ to $4$ groups and apply the self-attention 
on each group independently.
We merge all the updated feature map of each group together 
as the output $\mathbf{Z}^{\mathbf{L}}$.
For the short range attention, we permute the $\mathbf{Z}^{\mathbf{L}}$
to group the originally nearby 
positions together and get $\mathbf{X}^{\mathbf{S}}$.
Then we divide and apply self-attention following the same manner as long-range attention,
obtain the final feature map $\mathbf{Z}^{\mathbf{S}}$.
We can propagate the information from all the input positions to
each output position with the combination of the long-range attention 
and the short-range attention
(Best viewed in color).
}}
\vspace{-0.5cm}
\label{fig:interlaced_attn}
\end{figure*}

\paragraph{Interlacing/Interleaving.}
The previous works have applied the interlacing mechanism for network architecture design such as the Interleaved Group Convolution~\cite{igcv3,igcv2,igcv1}, ShuffleNet~\cite{Zhang_2018,ma2018shufflenet}
and the Channel Local Convolution~\cite{clcnet}.
Besides, the previous space-to-channel mechanism~\cite{Shi_2016_CVPR,Sajjadi_2018_CVPR,yang2019deeperlab} is also very similar
to the interlacing mechanism.
Our work is different from them as we mainly apply the interlacing mechanism
to decompose the dense affinity matrix within self-attention mechanism
with the product of two sparse affinity matrices,
and we apply the interlacing mechanism to group the pixels with long spatial interval 
distances together for the long-range attention.
Notably, the concurrent work Sparse Transformer~\cite{child2019generating} also applies similar factorization scheme to improve the efficiency of self-attention on sequential tasks while we are focused on semantic segmentation.
\section{Approach}

In this section, we first revisit the self-attention mechanism (Sec.\ref{self-attention})
and illustrate the details of the proposed
interlaced sparse self-attention scheme (Sec.\ref{sparse-self-attention}).
Last, we present the implementation of our approach with a short code
based on PyTorch, which is also very easy to be implemented
on other platforms (Sec.\ref{implement}).

\subsection{Self-Attention}\label{self-attention}
\label{sec:detail_ia}
The self-attention scheme is described as below,
\begin{align}
    \mathbf{A} &= \mathrm{Softmax} (\frac{\theta(\mathbf{X})\phi(\mathbf{X})^{\top}}{\sqrt{d}}), \label{eq:1}\\
    \mathbf{Z} &= \mathbf{A} g(\mathbf{X}), \label{eq:2}
\end{align}
where $\mathbf{X}$ is the input feature map, $\mathbf{A}$ is the dense affinity matrix,
and $\mathbf{Z}$ is the output feature map.
We set their shapes as $\mathbf{X}, \mathbf{Z} \in \mathbb{R}^{N\times C}$
and $\mathbf{A}\in \mathbb{R}^{N\times N}$(
$N$ is the number of pixels and
$C$ is the number of channel), and each element of $\mathbf{A}$
records the similarities between two positions.
The self-attention~\cite{vaswani2017attention} uses
two different transform functions $\theta$ and $\phi$ to
transform the input to lower dimensional space,
where $\theta(\mathbf{X}), \phi(\mathbf{X}) \in \mathbb{R}^{N\times \frac{C}{2}}$.
The inner product on the lower dimensional space is used to compute
the dense affinity matrix $\mathbf{A}$.
The scaling factor $d$ is used to
to solve the small gradient problem of softmax function 
according to~\cite{vaswani2017attention} and $d=\frac{C}{2}$.
Self-attention uses the function $g$ to learn a better embedding 
and $g(\mathbf{X}) \in \mathbb{R}^{N\times C}$.

\subsection{Interlaced Sparse Self-Attention}\label{sparse-self-attention}
\label{sec:detail_ia}
The key spirit of the proposed interlaced sparse self-attention is to decompose the
dense affinity matrix $\mathbf{A}$ in the self-attention scheme with the product of
two sparse block affinity matrices including $\mathbf{A}^\mathrm{L}$ and $\mathbf{A}^\mathrm{S}$.
We illustrate how to estimate the $\mathbf{A}^\mathrm{L}$ with the long-range attention
and the $\mathbf{A}^\mathrm{S}$ with the short-range attention.
With the combination of the long-range attention and the short-range attention,
we can propagate information from all the input positions 
to each output position.
We illustrate our approach with 
an example in Figure~\ref{fig:interlaced_attn}.

\vspace{.1cm}
\noindent\textbf{Long-range Attention}.
The main point of the long-range attention is to apply the self-attention
on the subsets of positions that have long spatial interval distances.

As illustrated in Figure~\ref{fig:interlaced_attn},
we first apply a permutation 
on the input feature map $\mathbf{X}$ to compute $\mathbf{X}^\mathrm{L} = \operatorname{Permute}(\mathbf{X})$.
Then, we divide the $\mathbf{X}^\mathrm{L}$ into $\mathcal{P}$ partitions
and each partition contains $\mathcal{Q}$ neighboring positions ($N=\mathcal{P} \times \mathcal{Q}$) following:
 $\mathbf{X}^\mathrm{L} = [{\mathbf{X}^\mathrm{L}_1}^{\top}, {\mathbf{X}^\mathrm{L}_2}^{\top}, \cdots, {\mathbf{X}^\mathrm{L}_\mathcal{P}}^{\top}]^{\top}$, 
where each $\mathbf{X}^\mathrm{L}_p$ is a subset of $\mathbf{X}^\mathrm{L}$ 
and its shape is $\mathbb{R}^{\mathcal{Q}\times C}$.
 We apply the self-attention on each $\mathbf{X}^\mathrm{L}_p$ independently
 as below,
\begin{align}
    \mathbf{A}_p^\mathrm{L} &= \mathrm{Softmax} (\frac{\theta(\mathbf{X}^\mathrm{L}_p)\phi(\mathbf{X}^\mathrm{L}_p)^{\top}}{\sqrt{d}}), \label{eq:6}\\
    \mathbf{Z}^\mathrm{L}_p &= \mathbf{A}_p^\mathrm{L} g(\mathbf{X}^\mathrm{L}_p), \label{eq:7}
\end{align}
where $\mathbf{A}_p^\mathrm{L} \in \mathbb{R}^{\mathcal{Q}\times \mathcal{Q}}$
is a small affinity matrix based on all the positions
from $\mathbf{X}_p^\mathrm{L}$ and 
$\mathbf{Z}^\mathrm{L}_p \in \mathbb{R}^{\mathcal{Q}\times c}$ is the updated representation
based on $\mathbf{X}^\mathrm{L}_p$.
All the other choices including $d$, $\theta$, $\phi$ and $g$ are the same as the self-attention scheme.

Last, we merge all the $\mathbf{Z}^\mathrm{L}_p$ from different groups
and get the output $\mathbf{Z}^\mathrm{L} = [{\mathbf{Z}^\mathrm{L}_1}^{\top}, {\mathbf{Z}^\mathrm{L}_2}^{\top}, \cdots, {\mathbf{Z}^\mathrm{L}_\mathcal{P}}^{\top}]^{\top}$.
We illustrate the actual affinity matrix of the long-range attention $\mathbf{A}^\mathrm{L}$ as below,
\begin{align}
\mathbf{A}^\mathrm{L} & = 
\begin{bmatrix}
    \mathbf{A}^\mathrm{L}_{1} & 0 & \cdots & 0 \\
    0 & \mathbf{A}^\mathrm{L}_{2} & \cdots & 0 \\
    \vdots     &\vdots      & \ddots & \vdots \\
    0 & 0 & \cdots & \mathbf{A}^\mathrm{L}_\mathcal{P}
\end{bmatrix},  
\end{align}
where we can see that only the affinity values in the diagonal blocks are non-zero.
We can also use $\mathbf{A}^\mathrm{L} = \operatorname{diag}(\mathbf{A}^\mathrm{L}_{1}, \mathbf{A}^\mathrm{L}_{2},\cdots, \mathbf{A}^\mathrm{L}_\mathcal{P})$ to represent
the sparse block matrix for convenience.

\begin{figure}[t]
\hspace{-0.2cm}
\includegraphics[width=.475\textwidth]{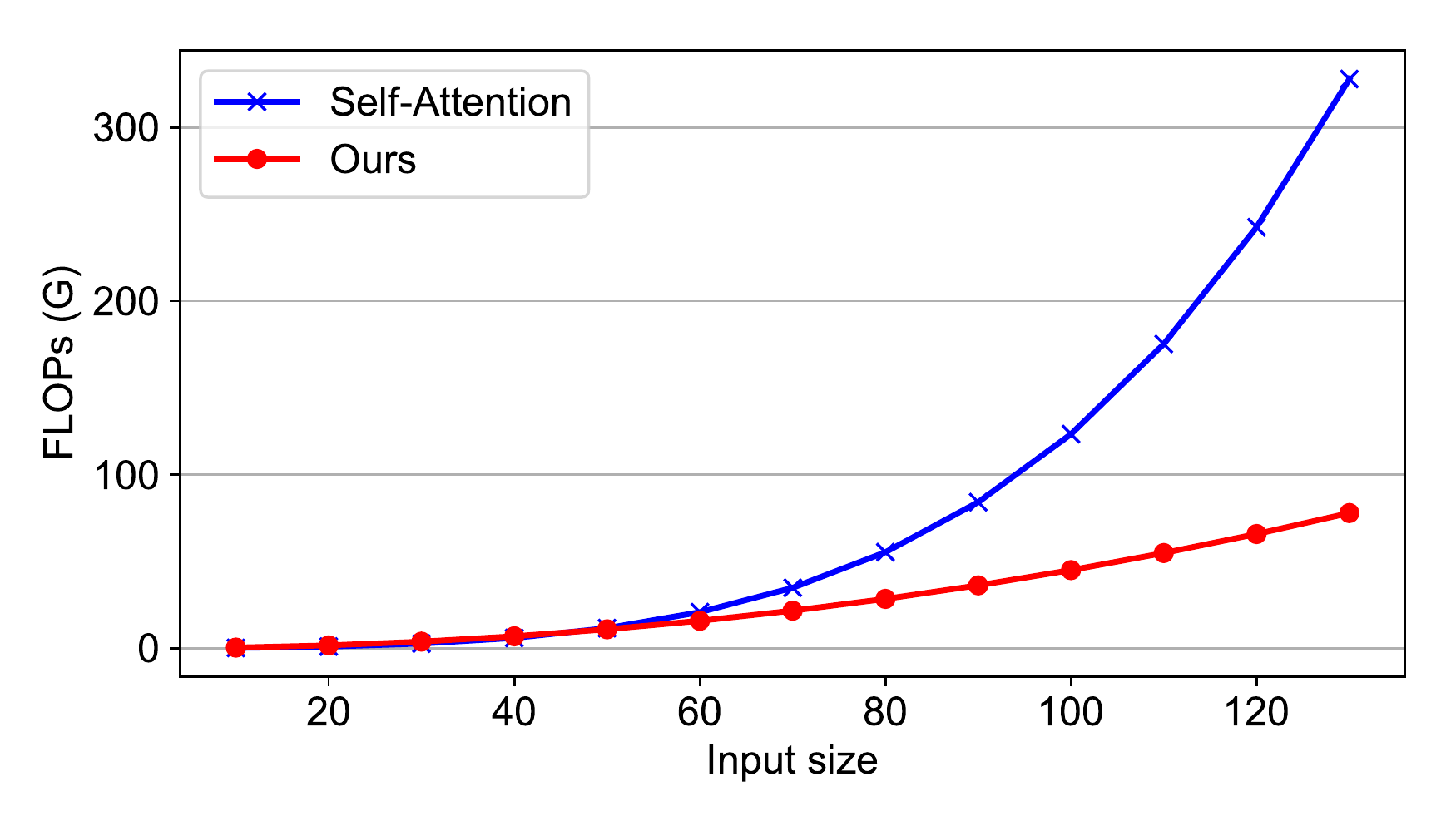}
\caption{\small{\textbf{FLOPs \emph{vs.} input size}. 
The $x$-axis represents the height or width of the input feature map (we assume
the height is equal to the width for convenience)
and the $y$-axis represents the computation cost measured with GFLOPs.
It can be seen the GFLOPs of self-attention mechanism
increases much faster than our approach with higher resolution inputs.
}}
\label{fig:complexity}
\vspace{-0.5cm}
\end{figure}

\vspace{.1cm}
\noindent\textbf{Short-range Attention}.
The main point of the short-range attention is to apply the self-attention
on the subsets of positions that have short spatial interval distances.

According to Figure~\ref{fig:interlaced_attn},
we apply another permutation on the output feature map from
the long-range attention following $\mathbf{X}^\mathrm{S} = \operatorname{Permute}(\mathbf{Z}^\mathrm{L})$.
Then, we divide $\mathbf{X}^\mathrm{S}$ into $\mathcal{Q}$ partitions
and each partition contains $\mathcal{P}$ neighboring positions following:
 $\mathbf{X}^\mathrm{S} = [{\mathbf{X}^\mathrm{S}_1}^{\top}, {\mathbf{X}^\mathrm{S}_2}^{\top}, \cdots, {\mathbf{X}^\mathrm{S}_\mathcal{Q}}^{\top}]^{\top}$, where each $\mathbf{X}^\mathrm{S}_q$ is of size $\mathbb{R}^{\mathcal{P}\times C}$.
 We apply the self-attention on each $\mathbf{X}^\mathrm{S}_q$ independently,
 which is similar with the Equation~\ref{eq:6} and Equation~\ref{eq:7}
 in the long-range attention.
 Accordingly, we can get $\mathbf{A}_q^\mathrm{S}$ and $\mathbf{Z}_q^\mathrm{S}$,
where $\mathbf{A}_q^\mathrm{S} \in \mathbb{R}^{\mathcal{P}\times \mathcal{P}}$
is a small affinity matrix based on $\mathbf{X}_q^\mathrm{S}$,
$\mathbf{Z}^\mathrm{S}_q \in \mathbb{R}^{\mathcal{P}\times C}$
is the updated representation based on $\mathbf{X}_q^\mathrm{S}$.

Last, we merge all the updated representation as
$\mathbf{Z}^\mathrm{S} = [{\mathbf{Z}^\mathrm{S}_1}^{\top}, {\mathbf{Z}^\mathrm{S}_2}^{\top}, \cdots, {\mathbf{Z}^\mathrm{S}_\mathcal{Q}}^{\top}]^{\top}$.
The sparse block affinity for
the short-range attention is illustrated as below,
\begin{align}
\mathbf{A}^\mathrm{S} & = \operatorname{diag}(\mathbf{A}^\mathrm{S}_{1}, \mathbf{A}^\mathrm{S}_{2},\cdots, \mathbf{A}^\mathrm{S}_\mathcal{Q}),
\end{align}
where we can see that the actual affinity matrix $\mathbf{A}^\mathrm{S}$
for the short-range attention is also very sparse and most of the affinity values are zero.

Finally, we directly fuse the output $\mathbf{Z}^\mathrm{S}$ 
from the short-range attention 
and the input feature map $\mathbf{X}$ following the previous works~\cite{wang2018non,OCNet}.
Especially, all of our above analysis can be generalized to higher dimensional inputs easily.

\vspace{.1cm}
\noindent\textbf{Complexity}.
Given an input feature map of size $H\times W  \times C$,
we analyze the computation / memory cost of both the self-attention mechanism
and our approach.

The complexity of self-attention mechanism is 
$\mathcal{O}(2HWC^2+\frac{3}{2}(HW)^2{C})$.
And the complexity of our approach is
$\mathcal{O}(4HWC^2+\frac{3}{2}(HW)^2{C}(\frac{1}{\mathcal{P}_{h}\mathcal{P}_{w}}+\frac{1}{\mathcal{Q}_{h}\mathcal{Q}_{w}}))$,
where we divide the height dimension to $\mathcal{P}_{h}$ groups
and the width dimension to $\mathcal{P}_{w}$ groups in the long-range attention
and $\mathcal{Q}_{h}$ and $\mathcal{Q}_{w}$ groups
during the short-range attention, 
and
$H = \mathcal{P}_{h}\mathcal{Q}_{h}$, $W = \mathcal{P}_{w}\mathcal{Q}_{w}$.
The complexity of our approach can be minimized to $\mathcal{O}(4HWC^2+3(HW)^{\frac{3}{2}}{C})$
when $\mathcal{P}_{h}\mathcal{P}_{w} = \sqrt{HW}$.
We compare the numerical complexity of our approach 
and the self-attention in Figure~\ref{fig:complexity}.
It can be seen that our approach is much more 
efficient than the conventional self-attention when processing inputs
of high resolution.

{
\lstset{
  backgroundcolor=\color{white},
  basicstyle=\fontsize{7.5pt}{8.5pt}\fontfamily{lmtt}\selectfont,
  columns=fullflexible,
  breaklines=true,
  captionpos=b,
  commentstyle=\fontsize{8pt}{9pt}\color{codegray},
  keywordstyle=\fontsize{8pt}{9pt}\color{codegreen},
  stringstyle=\fontsize{8pt}{9pt}\color{codeblue},
  frame=tb,
  otherkeywords = {self},
}
\begin{figure}[t]
\tiny
\begin{lstlisting}[language=python]
def InterlacedSparseSelfAttention(x, P_h, P_w):
    # x: input features with shape [N,C,H,W]
    # P_h, P_w: Number of partitions along H and W dimension

    N, C, H, W = x.size()
    Q_h, Q_w = H // P_h, W // P_w
    x = x.reshape(N, C, Q_h, P_h, Q_w, P_w)
    
    # Long-range Attention
    x = x.permute(0, 3, 5, 1, 2, 4)
    x = x.reshape(N * P_h * P_w, C, Q_h, Q_w)
    x = SelfAttention(x)
    x = x.reshape(N, P_h, P_w, C, Q_h, Q_w)
    
    # Short-range Attention
    x = x.permute(0, 4, 5, 3, 1, 2)
    x = x.reshape(N * Q_h * Q_w, C, P_h, P_w)
    x = SelfAttention(x)
    x = x.reshape(N, Q_h, Q_w, C, P_h, P_w)

    return x.permute(0, 3, 1, 4, 2, 5).reshape(N, C, H, W)
\end{lstlisting}
\caption{Python code of our approach based on PyTorch.}
\label{fig:code}
\vspace{-1em}
\end{figure}
}

\subsection{Implementation} \label{implement}
Our approach can be easily implemented with a few lines of code in Python.
We show the code of our approach in ~\ref{fig:code} based on PyTorch.
For the self-attention operation,
we directly use the open-source implementation from~\cite{OCNet}.
As illustrated in Figure~\ref{fig:code},
we implement the $\operatorname{Permutaion}$ and $\operatorname{Divide}$ operation 
in the long-range attention / short-range attention
by calling the default 
$\operatorname{permute}$ and $\operatorname{reshape}$ functions.
Besides, we implement all the transform functions including 
$\theta,\phi$ and $g$ with $1\times 1$ convolution + BN + ReLU.
The output channels of $\theta$ function and $\phi$ function are set as
half of the input channels, whereas the output channel of $g(\cdot)$
remains the same as the input channels.
Following~\cite{vaswani2017attention, wang2018non, OCNet},
we fuse the input feature map and the feature map output from 
our interlaced sparse self-attention via residual connection
or concatenation.

\section{Experiments}
We first compare our approach with the recent state-of-the-art
on six challenging semantic segmentation benchmarks including:
Cityscapes~\cite{cordts2016cityscapes}, ADE$20$K~\cite{zhou2017scene}, 
LIP~\cite{liang2018look}, PASCAL VOC $2012$~\cite{everingham2010pascal},
PASCAL-Context~\cite{mottaghi2014role}
and COCO-Stuff~\cite{caesar2018coco}.
Then we study the application of our method to the Mask R-CNN 
baseline~\cite{he2017mask} on the object detection and
instance segmentation benchmark COCO~\cite{lin2014microsoft}.
Finally, we conduct extensive ablation studies on our approach.

\subsection{Semantic Segmentation}
We first illustrate the details of all six benchmarks,
and then we provide the related results on each benchmark.
Especially,
we use mIoU (mean of class-wise intersection over union) 
and pixel accuracy as evaluation metrics on all six
semantic segmentation benchmarks.

\vspace{.1cm}
\noindent\textbf{Cityscapes}. 
The dataset
contains $5,000$ finely annotated images with $19$ semantic classes.
The images are in $2048\times 1024$ resolution
and captured from 50 different cities.
The training, validation, and test sets consist of 
$2,975$, $500$, $1,525$ images respectively.

\vspace{.1cm}
\noindent\textbf{ADE}$\textbf{20}$\textbf{K}. 
The dataset is very challenging
that contains $22$K densely annotated images
with $150$ fine-grained semantic concepts.
The training and validation sets consist of 
$20$K, $2$K images respectively.

\vspace{.1cm}
\noindent\textbf{LIP}.
The dataset is a large-scale dataset
that focuses on semantic understanding of human bodies.
It contains $50$K images with $19$ semantic human part labels
and $1$ background label for human parsing.
The training, validation, and test sets consist of 
$30$K, $10$K, $10$K images respectively.

\vspace{.1cm}
\noindent\textbf{PASCAL VOC $\textbf{2012}$}.
The dataset is a conventional object-centric segmentation dataset.
It contains more than $13$K images with 
$20$ object labels and $1$ background label.
The training, validation and test sets consist of about
$10$K, $1$K and $1$K images respectively.

\vspace{.1cm}
\noindent\textbf{PASCAL-Context}.
The dataset is a scene parsing dataset that contains $59$ semantic classes and $1$ background class.
The training and validation set consist of about $5$K and $5$K images respectively.

\vspace{.1cm}
\noindent\textbf{COCO-Stuff}.
The dataset is a very challenging segmentation dataset that involves $80$ object classes and $91$ stuff classes.
The training and validation set consist of $9$K and $1$K images respectively.

\vspace{.1cm}
\noindent\textbf{Network}.
We use ImageNet-pretrained ResNet-$50$/ResNet-$101$
as our backbone~\cite{long2015fully}.
Following the common practice~\cite{chen2017rethinking},
we remove the last two down-sample operation in the ResNet-$50$/ResNet-$101$
and employ dilated convolutions in last two stages, 
thus the size of the output feature map
is $8\times$ smaller than input image.
Following~\cite{OCNet, fu2018dual},
we reduce the number of channels of the output feature map
to $512$ with a $3\times3$ convolution.
Then we apply interlaced sparse self-attention module on the reduced feature map
and obtain a feature map $\mathbf{Z}$ of size $512\times H \times W$.
We directly predict (and upsample) the segmentation map based on $\mathbf{Z}$.


\vspace{.1cm}
\noindent\textbf{Training setting}.
For all the six semantic segmentation benchmarks,
we use the "poly" learning rate policy where the learning
rate is multiplied by $(1-(\frac{iter}{iter_{max}})^{power})$
with power as $0.9$.
We choose momentum of $0.9$ and a weight decay of $0.0005$.
Besides, 
we also apply an auxiliary loss
on the intermediate feature map after res-$4$ stage of ResNet
with a weight of $0.4$ following the PSPNet~\cite{zhao2017pyramid}.
For the data augmentation,
we apply random horizontal flip, random scaling (from $0.5$ to $2.0$)  
and random crop over all the training images.
Especially, we use the synchronized batch normalization~\cite{Bulò_2018_CVPR}
in all of our experiments.
We use $4\times$ P$100$ GPUs for the training of all of our experiments. 

We adopt different initial learning rates, batch sizes and training epochs by following
the previous works.
For  Cityscapes, we choose initial learning rate of $0.01$, batch size of $8$
and crop size of $769\times 769$~\cite{chen2017rethinking,zhao2017pyramid}.
For ADE$20$K, we choose initial learning rate of $0.02$, batch size of $16$ 
and crop size of $520\times 520$ following~\cite{zhao2017pyramid,OCNet}.
For LIP, we choose initial learning rate of $0.007$, batch size of $40$ 
and crop size of $473\times 473$ following ~\cite{liu2018devil}.
For PASCAL VOC $2012$ and PASCAL-Context, we choose initial learning rate of $0.01$,
batch size of $16$ and crop size of $513\times 513$
following~\cite{chen2018deeplab,Zhang_2018_CVPR}.
For COCO-Stuff, we choose initial learning rate of $0.01$,
batch size of $16$ and crop size of $520\times 520$.
We train the models for $110$ epochs on Cityscapes, $120$ epochs on ADE$20$K,
$150$ on LIP, $80$ epochs on PASCAL VOC $2012$, PASCAL-Context
and $100$ epochs on COCO-Stuff.

\vspace{.1cm}
\noindent\textbf{Results on Cityscapes}.
We report the results in Table~\ref{table:exp_cs_sota}
to compare our approach with
the recent state-of-the-arts on the test set of Cityscapes,
where we apply the multi-scale testing and flip testing
following the previous works.
Our approach outperforms most of other approaches when only using
the fine datasets for training.
For example, our approach achieves $80.3$\%
and outperforms the previous AAF~\cite{aaf2018} by $1.2$\%.
Compared with DANet~\cite{fu2018dual} (based on channel self-attention and spatial self-attention), our approach requires much smaller computation cost according to the complexity comparison in Table~\ref{table:efficiency_compare}.



\begin{table}[t]
\centering
\small
\caption{\small{Comparison with state-of-the-arts on the test set of Cityscapes.
We report both results trained with and without validation set.}}
\vspace{0.1cm}
\resizebox{\linewidth}{!}
{
\begin{tabular}{l|c|c|c}
\toprule[1pt]
Method & Backbone & Validation set & mIoU ($\%$)  \\
\toprule[1pt] 
PSPNet~\cite{zhao2017pyramid}   & ResNet-$101$ & \xmark & $78.4$ \\ 
PSANet~\cite{psanet}            & ResNet-$101$ & \xmark & $78.6$ \\ 
AAF~\cite{aaf2018}              & ResNet-$101$ & \xmark & ${79.1}$ \\ 
RefineNet~\cite{lin2017refinenet}   & ResNet-$101$ & \cmark & $73.6$ \\ 
DUC-HDC~\cite{wang2017understanding}& ResNet-$101$ & \cmark & $77.6$ \\ 
DSSPN~\cite{Liang_2018_CVPR}        & ResNet-$101$ & \cmark & $77.8$ \\ 
SAC~\cite{Zhang_2017_ICCV}          & ResNet-$101$ & \cmark & $78.1$ \\
DepthSeg~\cite{Kong_2018_CVPR}      & ResNet-$101$ & \cmark & $78.2$ \\
BiSeNet~\cite{yu2018bisenet}        & ResNet-$101$ & \cmark & $78.9$ \\
DFN~\cite{Yu_2018_CVPR}             & ResNet-$101$ & \cmark & $79.3$ \\ 
TKCN~\cite{wu2018treestructured}    & ResNet-$101$ & \cmark & $79.5$ \\
PSANet~\cite{psanet}                & ResNet-$101$ & \cmark & $80.1$ \\
DenseASPP~\cite{Yang_2018_CVPR}     & DenseNet-$161$ & \cmark & $80.6$\\
SVCNet~\cite{ding2019semantic}      & ResNet-$101$ & \cmark & $81.0$ \\
DANet~\cite{fu2018dual}             & ResNet-$101$ & \cmark & $\bf{81.5}$ \\
\hline
Ours                                & ResNet-$101$ & \xmark & $80.3$ \\ 
Ours                                & ResNet-$101$ & \cmark & $\underline{81.4}$ \\
\bottomrule[1pt] 
\end{tabular}
}
\label{table:exp_cs_sota}
\vspace{-0.2cm}
\end{table}

\vspace{.1cm}
\noindent\textbf{Results on ADE$20$K}.
In Table~\ref{table:ianet_sota_exp_ade20k},
we compare our approach with the state-of-the-arts
on the validation set of ADE$20$K.
For fair comparison, we employ ResNet-$101$ backbone 
and multi-scale testing following other methods.
From the results, we can see that our approach
achieves better performance compared to all other methods.
For example, our approach achieves $45.04$\% mIoU,
which improves the recent GCU~\cite{li2018beyond}
that using the same backbone by $0.2$\%.
Especially, improving $0.2$\% is not neglectable
considering the improvements on ADE$20$K
is very challenging.

\begin{table}[t]
\centering
\small
\caption{\small{Comparison with state-of-the-arts
on the validation set of ADE20K.}}
\vspace{0.1cm}
\begin{tabular}{l|c|c}
\toprule[1pt]
Method  & Backbone & mIoU ($\%$)  \\
\toprule[1pt]
RefineNet~\cite{lin2017refinenet}   & ResNet-$101$  &  $40.20$ \\ 
RefineNet~\cite{lin2017refinenet}   & ResNet-$152$  & $40.70$\\ 
UperNet~\cite{xiao2018unified}      & ResNet-$101$  &  $42.66$ \\
PSPNet~\cite{zhao2017pyramid}       & ResNet-$101$  &  $43.29$ \\ 
PSPNet~\cite{zhao2017pyramid}       & ResNet-$152$  &  $43.51$ \\ 
DSSPN~\cite{Liang_2018_CVPR}        & ResNet-$101$  &  $43.68$ \\ 
PSANet~\cite{psanet}                & ResNet-$101$  &  $43.77$ \\ 
SAC~\cite{Zhang_2017_ICCV}          & ResNet-$101$  &  $44.30$ \\
SGR~\cite{NIPS2018_7456}            & ResNet-$101$  &  $44.32$ \\
EncNet~\cite{Zhang_2018_CVPR}       & ResNet-$101$  &  $44.65$ \\
GCU~\cite{li2018beyond}             & ResNet-$101$  &  $\underline{44.81}$ \\
\hline
Ours                                & ResNet-$101$  & $\bf{45.04}$ \\
\bottomrule[1pt]
\end{tabular}
\label{table:ianet_sota_exp_ade20k}
\vspace{-0.2cm}
\end{table}

\vspace{.1cm}
\noindent\textbf{Results on LIP}.
We compare our approach to the previous state-of-the-arts on LIP
and illustrate the results in Table~\ref{table:ianet_sota_exp_lip}.
According to the results,
it can be seen that our approach achieves new state-of-the-art performance $55.07$\%,
which outperforms all the other methods using the same backbone by a large margin.
Notably, we only employ single scale testing following
CE2P~\cite{liu2018devil} and multi-scale testing
can be further incorporated to improve performance.


\begin{table}[t]
\centering
\small
\caption{\small{Comparison with state-of-the-arts
on the validation dataset of LIP.}}
\vspace{0.1cm}
\begin{tabular}{l|c|c}
\toprule[1pt]
Method  & Backbone & mIoU ($\%$)  \\
\toprule[1pt]
Attention+SSL~\cite{Gong_2017_CVPR} & ResNet-$101$  &  $44.73$\\ 
MMAN~\cite{luo2018macro}            & ResNet-$101$  &  $46.81$ \\ 
SS-NAN~\cite{Zhang_2017_ICCV}       & ResNet-$101$  &  $47.92$ \\ 
MuLA~\cite{nie2018mutual}           & ResNet-$101$  &  $49.30$ \\ 
JPPNet~\cite{liang2018look}         & ResNet-$101$  &  $51.37$ \\ 
CE2P~\cite{liu2018devil}            & ResNet-$101$  &  $\underline{53.10}$ \\ 
\hline
Ours                                & ResNet-$101$  & $\bf{55.07}$ \\
\bottomrule[1pt]
\end{tabular}
\label{table:ianet_sota_exp_lip}
\vspace{-0.3cm}
\end{table}

\begin{figure*}[htb]
\includegraphics[width=\textwidth]{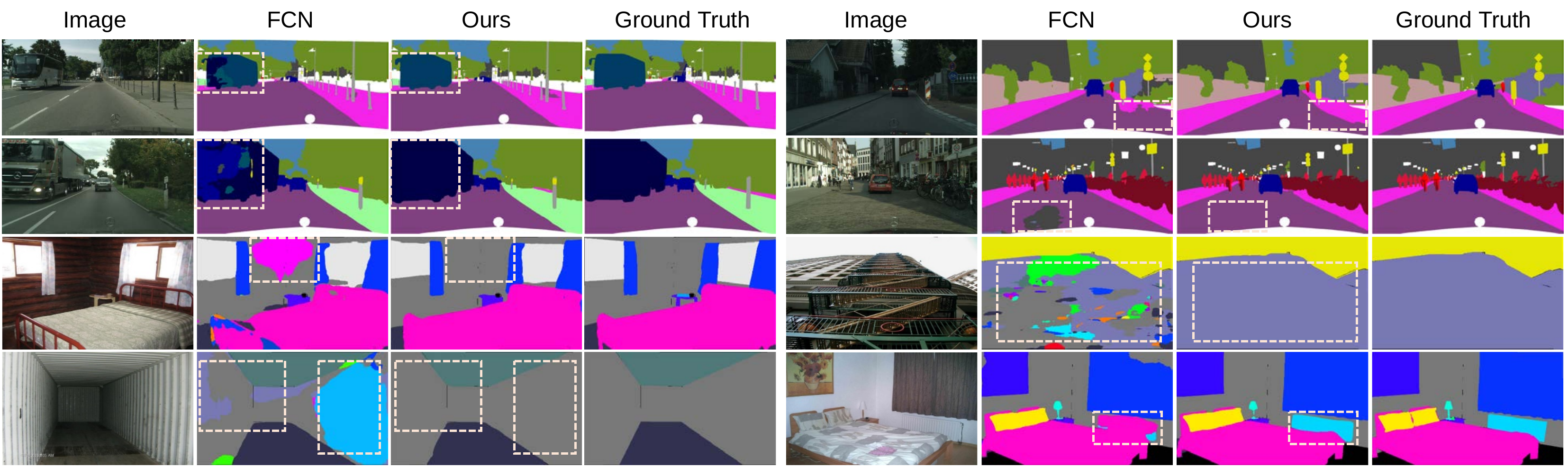}
\caption{\small{Visual improvements of our approach over the baseline on
both Cityscapes (first two rows) and ADE$20$K (last two rows).
We mark the improved regions with white dashed boxes
(Best viewed in color).
}}
\label{fig:vis}
\end{figure*}

\vspace{.1cm}
\noindent\textbf{Results on PASCAL VOC $\textbf{2012}$}.
The PASCAL VOC $2012$ dataset is one of gold standard benchmarks for semantic segmentation.
Following~\cite{Zhang_2018_CVPR,chen2017rethinking}, we first train the model on the \emph{trainaug} set
and then finetuned the model on the \emph{trainval} set.
As reported in Table~\ref{table:ianet_sota_exp_voc},
we achieve $83.2$\% mIoU on the PASCAL VOC $2012$ test set and slightly
outperforms the DANet while being much more efficient.

\begin{table}[t]
\centering
\small
\caption{\small{Comparison with state-of-the-arts
on the test set of PASCAL VOC $2012$ (w/o COCO data).}}
\vspace{0.1cm}
\begin{tabular}{l|c|c}
\toprule[1pt]
Method  & Backbone & mIoU ($\%$)  \\
\toprule[1pt]
FCN~\cite{long2015fully}            & VGG-16        &  $62.2$\\
DeepLab-CRF~\cite{chen2014semantic} & VGG-16        &  $71.6$\\
DPN~\cite{liu2015semantic}          & DPN           &  $74.1$\\
PSPNet~\cite{zhao2017pyramid}       & ResNet-$101$  &  $82.6$\\
DFN~\cite{Yu_2018_CVPR}             & ResNet-$101$  &  $82.7$\\
EncNet~\cite{Zhang_2018_CVPR}       & ResNet-$101$  &  $\underline{82.9}$\\
DANet~\cite{fu2018dual}             & ResNet-$101$  &  $82.6$\\
\hline
Ours                                & ResNet-$101$  & $\bf{83.2}$ \\
\bottomrule[1pt]
\end{tabular}
\label{table:ianet_sota_exp_voc}
\vspace{-0.3cm}
\end{table}

\vspace{.1cm}
\noindent\textbf{Results on PASCAL-Context}.
As illustrated in Table~\ref{table:ianet_sota_exp_context},
we compare our approach with the previous state-of-the-arts
on the validation set of PASCAL-Context.
Our approach achieves $54.1$\% mIoU and outperforms all the other
approaches.

\begin{table}[t]
\centering
\small
\caption{\small{Comparison with state-of-the-arts
on the validation set of PASCAL-Context.
mIoU is evaluated on 60 classes w/ background.}}
\vspace{0.1cm}
\begin{tabular}{l|c|c}
\toprule[1pt]
Method  & Backbone & mIoU ($\%$)  \\
\toprule[1pt]
DeepLabv2~\cite{chen2018deeplab}    & ResNet-$101$  &  $45.7$\\
UNet++~\cite{zhou2018unet++}        & ResNet-$101$  &  $47.7$\\
PSPNet~\cite{zhao2017pyramid}       & ResNet-$101$  &  $47.8$\\
CCL~\cite{ding2018ccl}              & ResNet-$101$  &  $51.6$\\
EncNet~\cite{Zhang_2018_CVPR}       & ResNet-$101$  &  $51.7$\\
DANet~\cite{fu2018dual}             & ResNet-$101$  &  $52.6$\\
SVCNet~\cite{ding2019semantic}      & ResNet-$101$  &  $\underline{53.2}$\\
\hline
Ours                                & ResNet-$101$  & $\bf{54.1}$ \\
\bottomrule[1pt]
\end{tabular}
\label{table:ianet_sota_exp_context}
\vspace{-0.3cm}
\end{table}


\vspace{.1cm}
\noindent\textbf{Results on COCO-Stuff}.
We further compare our method with the previous state-of-the-arts
on the validation set of COCO-Stuff benchmark.
According to the results illustrated in Table~\ref{table:ianet_sota_exp_coco_suff},
our method also achieves competitive performance while being more
efficient than the conventional self-attention based approaches.

\begin{table}[t]
\centering
\small
\caption{\small{Comparison with state-of-the-arts
on the validation set of COCO-Stuff.}}
\vspace{0.1cm}
\begin{tabular}{l|c|c}
\toprule[1pt]
Method  & Backbone & mIoU ($\%$)  \\
\toprule[1pt]
FCN~\cite{long2015fully}            & VGG-$16$      &  $22.7$\\
DAG-RNN~\cite{shuai2017dag}         & VGG-$16$      &  $31.2$\\
RefineNet~\cite{lin2017refinenet}   & ResNet-$101$  &  $33.6$\\
CCL~\cite{ding2018ccl}              & ResNet-$101$  &  $35.7$\\
SVCNet~\cite{ding2019semantic}      & ResNet-$101$  &  $39.6$\\
DANet~\cite{fu2018dual}             & ResNet-$101$  &  $\bf{39.7}$\\
\hline
Ours                                & ResNet-$101$  & $39.2$ \\
\bottomrule[1pt]
\end{tabular}
\label{table:ianet_sota_exp_coco_suff}
\vspace{-0.3cm}
\end{table}

\subsection{Application to Mask-RCNN}

\vspace{.1cm}
\noindent\textbf{Dataset.}
We use COCO~\cite{lin2014microsoft} dataset to evaluate our approach.
The dataset is one of the most challenging datasets
for object detection and instance segmentation,
which contains $140$K images annotated
with object bounding boxes and masks of 80 categories.
We follow the COCO2017 split as in~\cite{he2017mask},
where the training, validation and test sets contains
$115$K, $5$K, $20$K images respectively.
We report the standard COCO metrics including Average Precision (AP),
AP$_{50}$ and AP$_{75}$
for both bounding boxes and masks.

\vspace{.1cm}
\noindent\textbf{Training settings.}
We use Mask-RCNN~\cite{he2017mask} as baseline to conduct our experiments.
Similar to~\cite{wang2018non}, we insert $1$ 
non-local or interlaced sparse self-attention block
before the last block of res-$4$ stage of
the ResNet-$50$ FPN~\cite{lin2017fpn} backbone.
All models are initialized with ImageNet pretrained weights
and built upon open source toolbox~\cite{massa2018mrcnn}.
We train the models using SGD with batch size of $16$
and weight decay of $0.0001$.
We conduct experiments using training schedules including
`$1\times$ schedule' and `$2\times$ schedule'~\cite{massa2018mrcnn}. The $1\times$ schedule
starts at a learning rate of $0.02$ and is decreased by a factor of $10$
after $60$K and $80$K iterations and finally terminates at $90$K iterations.
We train for $180$K iterations for $2\times$ schedule and 
decreases the learning rate proportionally.
The other training and inference strategies keep the same
with the default settings in the~\cite{massa2018mrcnn}.

\vspace{.1cm}
\noindent\textbf{Results.}
We report the results on COCO dataset in Table~\ref{table:exp_coco}.
We can see that adding one non-local block~\cite{wang2018non} or interlaced sparse self-attention module 
consistently improves the Mask-RCNN baseline by $\sim1$\%
on all metrics involving both object detection and instance segmentation.
Similar gains are observed for both $1\times$ schedule and $2\times$ schedule.
For example, our approach improves the box AP/mask AP of Mask-RCNN
from $38.7$/$34.9$ to $39.7$/$35.7$ with $2\times$ schedule.
Especially, the performance of our approach is comparable
with the non-local block on all metrics 
while decreasing the computation complexity significantly.

\begin{table}[t]
\centering
\small
\caption{\small{Comparison with non-local~\cite{wang2018non} (NL)
on the validation set of COCO.
We use Mask-RCNN~\cite{he2017mask} as baseline
and employ ResNet-$50$ FPN backbone for all models.}}
\vspace{0.2cm}
\resizebox{\linewidth}{!}{
\begin{tabular}{l|c|ccc|ccc}
\toprule[1pt]
Method          & Schedule & $AP^{box}$ & $AP^{box}_{50}$ & $AP^{box}_{75}$ & $AP^{mask}$ & $AP^{mask}_{50}$  & $AP^{box}_{75}$\\
\toprule[1pt]
Mask-RCNN   & $1\times$ & $37.7$ & $59.2$ & $41.0$ & $34.2$ & $56.0$ & $36.2$ \\ 
+ NL        & $1\times$ & $\bf{38.8}$ & $60.6$ & $42.3$ & $35.1$ & $\bf{57.4}$ & $37.3$ \\
+ Ours      & $1\times$ & $\bf{38.8}$ & $\bf{60.7}$ & $\bf{42.5}$ & $\bf{35.2}$ & $57.3$ & $\bf{37.6}$\\
\hline
Mask-RCNN   & $2\times$ & $38.7$ & $59.9$ & $42.1$ & $34.9$ & $56.8$ & $37.0$ \\ 
+ NL        & $2\times$ & $\bf{39.7}$ & $\bf{61.3}$ & $\bf{43.4}$ & $\bf{35.9}$ & $\bf{58.3}$ & $\bf{38.2}$ \\
+ Ours      & $2\times$ & $\bf{39.7}$ & $61.1$ & $43.3$ & $35.7$ & $57.8$ & $38.1$ \\
\bottomrule[1pt]
\end{tabular}
}
\label{table:exp_coco}
\end{table}

\begin{figure*}[t]
\includegraphics[width=\textwidth]{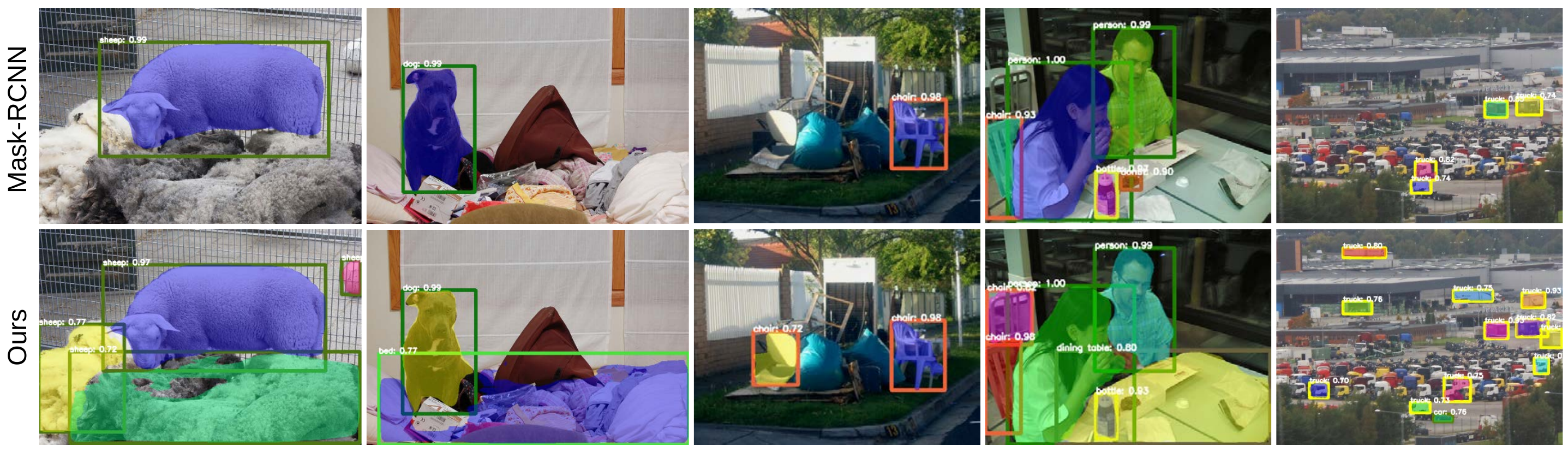}
\caption{\small{Visualization of the object detection and instance segmentation results
of Mask-RCNN~\cite{he2017mask} and our approach on the validation set of COCO (Best viewed in color).
}}
\vspace{-0.3cm}
\label{fig:vis_coco}
\end{figure*}

\subsection{Ablation Study}

\begin{table}[t]
\centering
\small
\caption{\small{Influence of $\mathcal{P}_h$ and $\mathcal{P}_w$ within
the interlaced sparse self-attention on the validation set of Cityscapes.}}
\vspace{0.1cm}
\begin{tabular}{l|c|c|c|c}
\toprule[1pt]
 Method & $\mathcal{P}_h$ & $\mathcal{P}_w$ &Pixel Acc ($\%$) & mIoU ($\%$)  \\ \hline
\toprule[1pt]
Baseline    & - & -  & $96.08$ & $75.90$ \\ \hline
\multirow{7}{*}{+ Ours} &  $4$ & $4$      & $96.30$ & $78.97$ \\ 
 & $4$ & $8$      & $96.31$ & $78.95$ \\ 
 & $8$ & $4$      & $96.32$ & $79.31$ \\ 
 & $8$ & $8$      & $\bf{96.33}$ & $\bf{79.49}$ \\ 
 & $8$ & $16$     & $96.29$ & $79.04$ \\ 
 & $16$ & $8$     & $96.19$ & $78.90$ \\ 
 & $16$ & $16$    & $96.32$ & $79.40$ \\ 
\bottomrule[1pt] 
\end{tabular}
\label{table:ia_factor}
\vspace{-0.3cm}
\end{table}

\vspace{.1cm}
\noindent\textbf{Influence of the partition numbers.}
We investigate the influence of the partition numbers of our approach,
e.g., $\mathcal{P}$, $\mathcal{Q}$.
We conduct the experiments with various choices of $\mathcal{P}$ and $\mathcal{Q}$,
and present the related results in Table~\ref{table:ia_factor}.
Especially, we can determine the value of 
$\mathcal{Q}_{h}$ and $\mathcal{Q}_{w}$
according to $\mathcal{P}_{h}$ and $\mathcal{P}_{w}$.
According to the results in Table~\ref{table:ia_factor},
it can be seen
that our approach consistently improves over the
baseline under various choices of the hyper-parameters.
And the choice $\mathcal{P}_{h}=\mathcal{P}_{w}=8$
achieves slightly better results than others.
We simply set $\mathcal{P}_{h}=\mathcal{P}_{w}=8$ for all of our experiments
for convenience.
All the above experiments use the ResNet-$101$ FCN
as the backbone and use the same training and testing settings.

\vspace{.1cm}
\noindent\textbf{Long+Short \emph{vs.} Short+Long.}
Considering that we can achieve the fully dense information propagation
with either long-range attention first or short-range attention first,
we study the influence of the order of the long-range attention
and the short-range attention on the semantic segmentation task.
We simply conduct another group of experiments to firstly apply 
the short-range attention and then the long-range-attention.
The related results are summarized in Table~\ref{table:exp_ablation_order}.
It can be seen that changing the order still
improves over the baseline for a large margin.
Applying the long-range attention firstly seems to be prefered for the
semantic segmentation tasks.
We use the ResNet-$101$ FCN
as the backbone for all the above experiments.
In all of our experiments, we apply the long-range attention firstly
unless otherwise specified.

\begin{table}[t]
\centering
\small
\caption{\small{Impact of the order of long-range and short-range attention
on the validation set of Cityscapes.}}
\vspace{0.1cm}
\begin{tabular}{l|c|c}
\toprule[1pt]
Method  & Pixel Acc ($\%$) & mIoU ($\%$) \\
\toprule[1pt]
Baseline      & $96.08$ & $75.90$ \\ \hline
+ Ours (Short + Long)            & $96.26$ & $79.10$ \\
+ Ours (Long + Short)            & $\bf{96.33}$ & $\bf{79.49}$ \\
\bottomrule[1pt]
\end{tabular}
\label{table:exp_ablation_order}
\vspace{-0.3cm}
\end{table}

\noindent\textbf{Comparison with Self-Attention/Non-local.}
We compare our approach with the self-attention/non-local mechanism
on three semantic segmentation datasets.
We report the related results in Table~\ref{table:compare-sa}.
We use ResNet-$50$ as the backbone for experiments on ADE$20$K
and ResNet-$101$ as the backbone for all the other experiments.
We can see that our approach outperforms the self-attention 
on all three datasets.
For example, our approach outperforms the self-attention based approach
by $0.56\%$/$0.96\%$/$0.84\%$ on ADE$20$K/Cityscapes/LIP respectively.

\noindent\textbf{Comparison with Pyramid Pooling Module.}
We also compare our approach with the well verified
Pyramid Pooling Module (PPM)~\cite{zhao2017pyramid}
under the same training/testing settings.
We report the related results in Table~\ref{table:compare-sa}.
We can see that our approach consistently outperforms the PPM on all three datasets.
Specifically, our approach outperforms the PPM
by $0.61\%$/$0.99\%$/$0.89\%$ on ADE$20$K/Cityscapes/LIP respectively.
In fact,
the advantage of self-attention over PPM
has also been convincingly verified 
in the previous works including DANet~\cite{fu2018dual}, CFNet~\cite{zhang2019co},
and OCNet~\cite{OCNet}. 
The (further) advantage of our approach 
compared with~\cite{fu2018dual,zhang2019co,OCNet} is the significant efficiency improvement.

\begin{table}[t]
\centering
\small
\caption{\small{Comparison with
self-attention (SA) and Pyramid Pooling Module (PPM) on three datasets.
We use CE$2$P~\cite{liu2018devil} as our baseline for LIP
and ResNet based FCN for Cityscapes and ADE$20$K.}}
\vspace{0.1cm}
\resizebox{\linewidth}{!}
{
\begin{tabular}{l|c|c|c|c} 
\toprule[1pt] 
Method           & Dataset & Backbone      & Pixel Acc ($\%$)  & mIoU ($\%$)  \\
\toprule[1pt]
Baseline & \multirow{4}{*}{ADE$20$K} & \multirow{4}{*}{ResNet-$50$} & $76.41$ & $34.35$ \\ 
+ PPM         &  &  & $80.17$            & $41.50$ \\ 
+ SA          &  &  & $80.19$           & $41.55$ \\
+ Ours        &  &  & $\bf{80.27}$  & $\bf{42.11}$ \\
\hline 
Baseline & \multirow{4}{*}{Cityscapes} & \multirow{4}{*}{ResNet-$101$} &$96.08$ & $75.90$\\
+ PPM         &  &  & $96.20$        & $78.50$ \\
+ SA          &  &  & $96.21$    & $78.53$ \\
+ Ours        &  &  & $\bf{96.33}$  & $\bf{79.49}$  \\ 
\hline 
CE$2$P        & \multirow{4}{*}{LIP} & \multirow{4}{*}{ResNet-$101$} & $87.37$ & $53.10$ \\
+ PPM         &  &    & $87.21$           & $54.18$ \\
+ SA          &  &    & $87.23$           & $54.23$ \\
+ Ours        &  &    & $\bf{87.69}$      & $\bf{55.07}$ \\
\bottomrule[1pt]
\end{tabular}
}
\label{table:compare-sa}
\end{table}



\noindent\textbf{Comparison with CGNL.}
The recently proposed compact generalized non-local
module (CGNL)~\cite{yue2018cgnl} also attempts to improve the efficiency of the original non-local mechanism.
Considering the CGNL is extensively evaluated
on the fine-grained classification dataset,
we compare our approach with CGNL on
CUB-$200$-$2011$~\cite{WahCUB_200_2011} dataset.

We use ResNet-$50$ as backbone to for all the experiments.
Following~\cite{wang2018non,yue2018cgnl},
we compare our approach with CGNL under two different settings:
(1) insert one CGNL or interlaced sparse self-attention module (to res-$4$);
(2) insert $5$ CGNL or interlaced sparse self-attention modules
($2$ modules to res-$3$ and $3$ modules to res-$4$).
We report the Top-$1$ and Top-$5$ classiﬁcation accuracy 
on the validation set of CUB-$200$-$2011$ in Table~\ref{table:exp_cub}.
It can be seen that our approach outperforms the previous 
CGNL under both kinds of settings.
Besides, we also verify the advantages of our approach over CGNL on a segmentation dataset (Cityscapes) and illustrate the performance in Table~\ref{table:exp_ablation_down}.

\begin{table}[t]
\centering
\small
\caption{\small{Comparison with CGNL~\cite{yue2018cgnl}
on the validation set of CUB-$200$-$2011$.}}
\vspace{0.1cm}
\begin{tabular}{l|c|c}
\toprule[1pt]
Method                      & Top$1$ Acc ($\%$) & Top$5$ Acc ($\%$) \\ 
\toprule[1pt]
ResNet-$50$                           & $84.37$  & $96.53$ \\ \hline 
+ $1\times$ CGNL~\cite{yue2018cgnl}   & $85.14$  & $96.88$ \\ 
+ $5\times$ CGNL~\cite{yue2018cgnl}   & $85.68$  & $96.69$ \\ 
\hline
+ $1\times$ CGNL (Our impl.)          & $85.69$            & $96.95$ \\
+ $5\times$ CGNL (Our impl.)          & $86.31$            & $97.05$ \\
+ $1\times$ Ours          & $86.28$            & $97.03$ \\
+ $5\times$ Ours          & $\bf{86.54}$       & $\bf{97.10}$ \\
\bottomrule[1pt]
\end{tabular}
\label{table:exp_cub}
\end{table}

\vspace{.1cm}
\noindent\textbf{Comparison with Down-Sampling/RCCA.}
We compare our approach with other two kinds of mechanisms
that are conventionally used to improve the efficiency of self-attention/non-local
such as the down-sampling scheme~\cite{wang2018non}
and RCCA~\cite{huang2018ccnet}.
For the down-sampling scheme, we directly down-sample
the feature map for $2\times$ before computing the dense 
affinity matrix.
We evaluate both kinds of mechanisms on the 
validation set of Cityscapes and report 
the related performance
in Table~\ref{table:exp_ablation_down}.
It can be seen that our approach 
outperforms the above two mechanisms
according to the results in Table~\ref{table:exp_ablation_down}.
Especially, we use the average performance ($79.12\%$)
reported in~\cite{huang2018ccnet} for fairness.



\begin{table}[t]
\centering
\small
\caption{\small{Comparison with
SA~\cite{vaswani2017attention} with $2\times$ downsampling,
RCCA~\cite{huang2018ccnet} and CGNL~\cite{yue2018cgnl}
on the validation set of Cityscapes.
All methods choose ResNet-$101$ FCN as the backbone.}}
\vspace{0.1cm}
\begin{tabular}{l|c|c}
\toprule[1pt]
Method                      & Pixel Acc ($\%$) & mIoU ($\%$)  \\
\toprule[1pt]
Baseline                    & $96.08$ & $75.90$ \\ \hline
+ RCCA~\cite{huang2018ccnet}& - & $79.12$       \\ 
\hline
+ SA-$2\times$ (Our impl.)  & $96.12$ & $76.49$ \\ 
+ RCCA (Our impl.)          & $96.28$ & $78.63$ \\ 
+ CGNL (Our impl.)          & $96.01$ & $77.01$     \\ 
+ Ours &                    $\bf{96.33}$ & $\bf{79.49}$ \\ 
\bottomrule[1pt]
\end{tabular}
\label{table:exp_ablation_down}
\vspace{-0.3cm}
\end{table}

\vspace{.1cm}
\noindent\textbf{Efficiency Comparison.}
We compare our approach with PPM~\cite{zhao2017pyramid},
SA~\cite{vaswani2017attention}, DANet~\cite{fu2018dual}, 
RCCA~\cite{huang2018ccnet} and CGNL~\cite{yue2018cgnl}
in terms of efficiency in this section.
We report the GPU memory, GFLOPs and inference time when processing
input feature map of size $2048\times128\times128$ in Table~\ref{table:efficiency_compare}.
It can be seen that 
our approach is much more efficient
than all the other approaches.

\begin{table}[t]
\centering
\small
\caption{\small{Efficiency comparison
given input feature map of size  [$2048\times128\times128$]
in inference stage.}}
\vspace{0.1cm}
\begin{tabular}{l|c|c|c}
\toprule[1pt]
Method              & Memory (MB)   & GFLOPs    &  Time (ms) \\
\toprule[1pt]
PPM~\cite{zhao2017pyramid}      & $664$         & $619$     & $75$  \\
SA~\cite{vaswani2017attention}  & $2168$        & $619$     & $77$  \\
DANet~\cite{fu2018dual}       & $2339$        & $1110$    & $121$  \\
RCCA~\cite{huang2018ccnet}      & $427$         & $804$     & $131$  \\
CGNL~\cite{yue2018cgnl}         & $266$         & $412$     & $75$  \\
\hline
Ours                & $\bf{252}$    & $\bf{386}$& $\bf{45}$  \\
\bottomrule[1pt]
\end{tabular}
\label{table:efficiency_compare}
\end{table}

\begin{figure}[t]
\includegraphics[width=.475\textwidth]{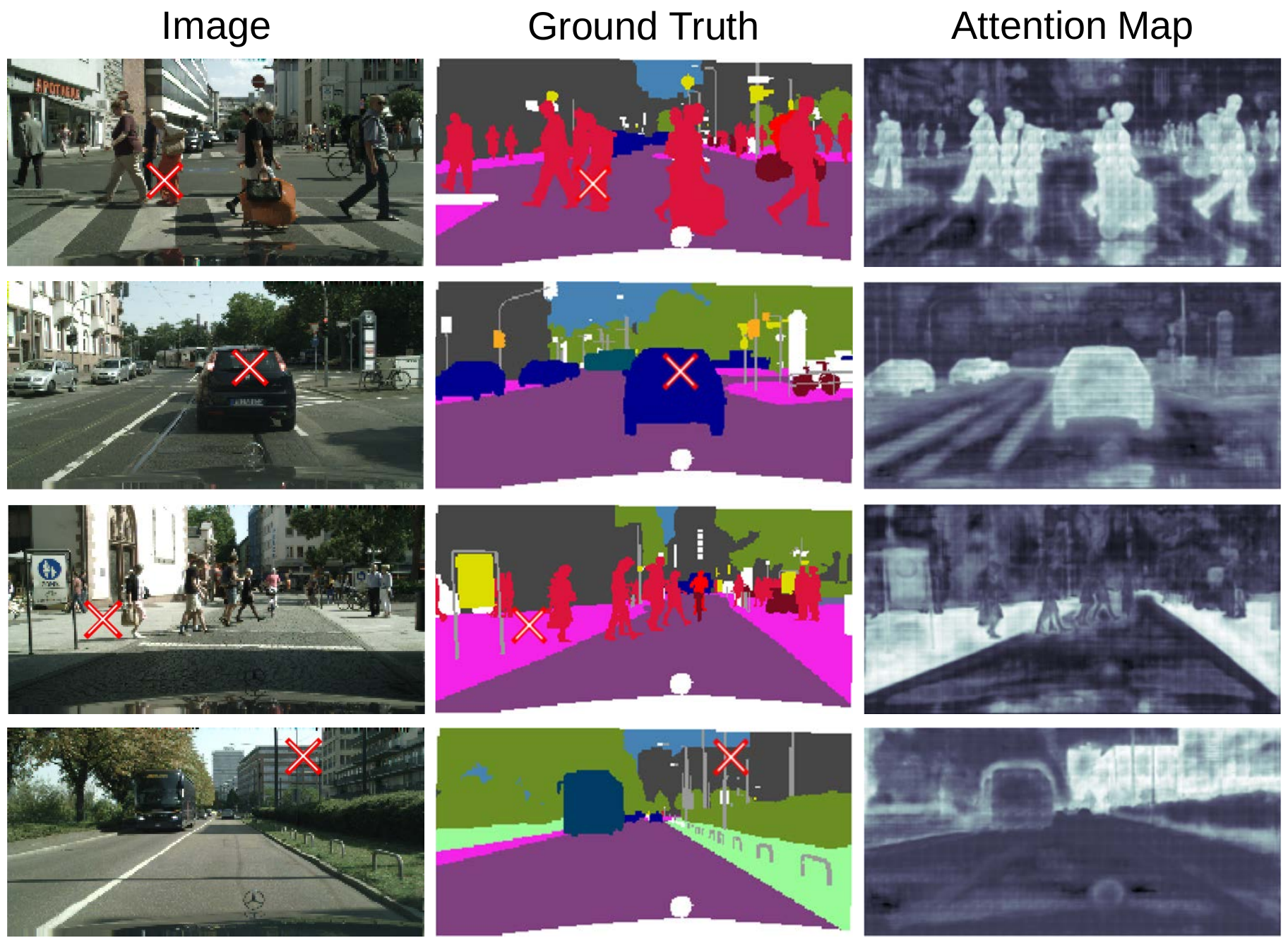}
\caption{\small{Visualization of attention maps learned with our approach on the validation set of Cityscapes.
We present an input image, the ground truth label map and attention map for the 
pixel marked by red cross in each row.
}}
\label{fig:vis_attn}
\vspace{-0.5cm}
\end{figure}

\vspace{.1cm}
\noindent\textbf{Visualization.}
First,
we visualize the segmentation maps
predicted with our approach and the baseline in Figure~\ref{fig:vis}.
The first two rows show the examples from the validation set of Cityscapes
and the other two rows show the examples from the validation set of ADE$20$K.
It can be seen that our method produces better segmentation maps
compared with the baseline.
We mark all the improved regions with white dashed boxes.

Second,
inspired by the recent~\cite{OCNet},
we also visualize the attention maps learned
with our approach
on the validation set of Cityscapes in Figure~\ref{fig:vis_attn}.
For each row,
we randomly select a pixel (marked by red cross)
in the image and visualize its attention map learned with our approach.
It can be seen that our approach also pays more attention
to the pixels that belong to same category as the chosen pixel,
which is similar to the observation of~\cite{OCNet}.

Last,
we visualize the object detection and instance segmentation results
of our approach and the Mask-RCNN
on the validation set of COCO in Figure~\ref{fig:vis_coco}.
It can be seen that our approach improves the Mask-RCNN consistently
on all the examples.
For example, the Mask-RCNN fails to detect multiple cars
in the last example while our approach achieves better 
detection performance.

\section{Conclusion}
In this work,
we have presented 
the interlaced sparse self-attention mechanism
to improve the efficiency
of self-attention scheme.
The main idea of our approach 
is very clear and simple:
factoring the dense affinty matrix
as the product of two
sparse affinity matrices.
Especially,
our approach is easy to implement
based on the existing implementation for
self-attention scheme.
We empirically compare our approach with various
existing approaches and show that our approach
achieves competitive performance on various semantic
segmentation datasets while being much more efficient
compared with the conventional self-attention mechanism.

{\small
\bibliographystyle{ieee}
\bibliography{egbib}

\begin{thebibliography}{10}\itemsep=-1pt

\bibitem{wiki_interlace}
\url{https://en.wikipedia.org/wiki/Interlacing_(bitmaps)}.

\bibitem{caesar2018coco}
H.~Caesar, J.~Uijlings, and V.~Ferrari.
\newblock Coco-stuff: Thing and stuff classes in context.
\newblock In {\em CVPR}, 2018.

\bibitem{chen2014semantic}
L.-C. Chen, G.~Papandreou, I.~Kokkinos, K.~Murphy, and A.~L. Yuille.
\newblock Semantic image segmentation with deep convolutional nets and fully
  connected crfs.
\newblock In {\em ICLR}, 2015.

\bibitem{chen2018deeplab}
L.-C. Chen, G.~Papandreou, I.~Kokkinos, K.~Murphy, and A.~L. Yuille.
\newblock Deeplab: Semantic image segmentation with deep convolutional nets,
  atrous convolution, and fully connected crfs.
\newblock {\em PAMI}, 2018.

\bibitem{chen2017rethinking}
L.-C. Chen, G.~Papandreou, F.~Schroff, and H.~Adam.
\newblock Rethinking atrous convolution for semantic image segmentation.
\newblock {\em arXiv:1706.05587}, 2017.

\bibitem{A2Net}
Y.~Chen, Y.~Kalantidis, J.~Li, S.~Yan, and J.~Feng.
\newblock A\^{2}-nets: Double attention networks.
\newblock In {\em NIPS}, 2018.

\bibitem{child2019generating}
R.~Child, S.~Gray, A.~Radford, and I.~Sutskever.
\newblock Generating long sequences with sparse transformers.
\newblock {\em arXiv:1904.10509}, 2019.

\bibitem{cordts2016cityscapes}
M.~Cordts, M.~Omran, S.~Ramos, T.~Rehfeld, M.~Enzweiler, R.~Benenson,
  U.~Franke, S.~Roth, and B.~Schiele.
\newblock The cityscapes dataset for semantic urban scene understanding.
\newblock In {\em CVPR}, 2016.

\bibitem{ding2018ccl}
H.~Ding, X.~Jiang, B.~Shuai, A.~Q. Liu, and G.~Wang.
\newblock Context contrasted feature and gated multi-scale aggregation for
  scene segmentation.
\newblock In {\em CVPR}, 2018.

\bibitem{ding2019semantic}
H.~Ding, X.~Jiang, B.~Shuai, A.~Q. Liu, and G.~Wang.
\newblock Semantic correlation promoted shape-variant context for segmentation.
\newblock In {\em CVPR}, 2019.

\bibitem{everingham2010pascal}
M.~Everingham, L.~Van~Gool, C.~K. Williams, J.~Winn, and A.~Zisserman.
\newblock The pascal visual object classes (voc) challenge.
\newblock {\em IJCV}, 2010.

\bibitem{fu2018dual}
J.~Fu, J.~Liu, H.~Tian, Z.~Fang, and H.~Lu.
\newblock Dual attention network for scene segmentation.
\newblock In {\em CVPR}, 2019.

\bibitem{Gong_2017_CVPR}
K.~Gong, X.~Liang, D.~Zhang, X.~Shen, and L.~Lin.
\newblock Look into person: Self-supervised structure-sensitive learning and a
  new benchmark for human parsing.
\newblock In {\em CVPR}, 2017.

\bibitem{he2017mask}
K.~He, G.~Gkioxari, P.~Doll{\'{a}}r, and R.~B. Girshick.
\newblock Mask {R-CNN}.
\newblock In {\em ICCV}, 2017.

\bibitem{hu2018relation}
H.~Hu, J.~Gu, Z.~Zhang, J.~Dai, and Y.~Wei.
\newblock Relation networks for object detection.
\newblock In {\em CVPR}, 2018.

\bibitem{huang2018ccnet}
Z.~Huang, X.~Wang, L.~Huang, C.~Huang, Y.~Wei, and W.~Liu.
\newblock Ccnet: Criss-cross attention for semantic segmentation.
\newblock In {\em ICCV}, 2019.

\bibitem{aaf2018}
T.-W. Ke, J.-J. Hwang, Z.~Liu, and S.~X. Yu.
\newblock Adaptive affinity fields for semantic segmentation.
\newblock In {\em ECCV}, 2018.

\bibitem{Kong_2018_CVPR}
S.~Kong and C.~C. Fowlkes.
\newblock Recurrent scene parsing with perspective understanding in the loop.
\newblock In {\em CVPR}, 2018.

\bibitem{li2018beyond}
Y.~Li and A.~Gupta.
\newblock Beyond grids: Learning graph representations for visual recognition.
\newblock In {\em NIPS}, 2018.

\bibitem{liang2018look}
X.~Liang, K.~Gong, X.~Shen, and L.~Lin.
\newblock Look into person: Joint body parsing \& pose estimation network and a
  new benchmark.
\newblock {\em PAMI}, 2018.

\bibitem{NIPS2018_7456}
X.~Liang, Z.~Hu, H.~Zhang, L.~Lin, and E.~P. Xing.
\newblock Symbolic graph reasoning meets convolutions.
\newblock In {\em NIPS}, 2018.

\bibitem{Liang_2018_CVPR}
X.~Liang, H.~Zhou, and E.~Xing.
\newblock Dynamic-structured semantic propagation network.
\newblock In {\em CVPR}, 2018.

\bibitem{liao2018video}
X.~Liao, L.~He, and Z.~Yang.
\newblock Video-based person re-identification via 3d convolutional networks
  and non-local attention.
\newblock {\em arXiv:1807.05073}, 2018.

\bibitem{lin2017refinenet}
G.~Lin, A.~Milan, C.~Shen, and I.~D. Reid.
\newblock Refinenet: Multi-path refinement networks for high-resolution
  semantic segmentation.
\newblock In {\em CVPR}, 2017.

\bibitem{lin2017fpn}
T.~Lin, P.~Doll{\'{a}}r, R.~B. Girshick, K.~He, B.~Hariharan, and S.~J.
  Belongie.
\newblock Feature pyramid networks for object detection.
\newblock In {\em CVPR}, 2017.

\bibitem{lin2014microsoft}
T.-Y. Lin, M.~Maire, S.~Belongie, J.~Hays, P.~Perona, D.~Ramanan,
  P.~Doll{\'a}r, and C.~L. Zitnick.
\newblock Microsoft coco: Common objects in context.
\newblock In {\em ECCV}, 2014.

\bibitem{liu2018devil}
T.~Liu, T.~Ruan, Z.~Huang, Y.~Wei, S.~Wei, Y.~Zhao, and T.~Huang.
\newblock Devil in the details: Towards accurate single and multiple human
  parsing.
\newblock {\em arXiv:1809.05996}, 2018.

\bibitem{liu2015semantic}
Z.~Liu, X.~Li, P.~Luo, C.-C. Loy, and X.~Tang.
\newblock Semantic image segmentation via deep parsing network.
\newblock In {\em ICCV}, 2015.

\bibitem{long2015fully}
J.~Long, E.~Shelhamer, and T.~Darrell.
\newblock Fully convolutional networks for semantic segmentation.
\newblock In {\em CVPR}, 2015.

\bibitem{luo2018macro}
Y.~Luo, Z.~Zheng, L.~Zheng, G.~Tao, Y.~Junqing, and Y.~Yang.
\newblock Macro-micro adversarial network for human parsing.
\newblock In {\em ECCV}, 2018.

\bibitem{ma2018shufflenet}
N.~Ma, X.~Zhang, H.-T. Zheng, and J.~Sun.
\newblock Shufflenet v2: Practical guidelines for efficient cnn architecture
  design.
\newblock In {\em ECCV}, 2018.

\bibitem{massa2018mrcnn}
F.~Massa and R.~Girshick.
\newblock {maskrcnn-benchmark: Fast, modular reference implementation of
  Instance Segmentation and Object Detection algorithms in PyTorch}.
\newblock \url{https://github.com/facebookresearch/maskrcnn-benchmark}, 2018.

\bibitem{mottaghi2014role}
R.~Mottaghi, X.~Chen, X.~Liu, N.-G. Cho, S.-W. Lee, S.~Fidler, R.~Urtasun, and
  A.~Yuille.
\newblock The role of context for object detection and semantic segmentation in
  the wild.
\newblock In {\em CVPR}, 2014.

\bibitem{nie2018mutual}
X.~Nie, J.~Feng, and S.~Yan.
\newblock Mutual learning to adapt for joint human parsing and pose estimation.
\newblock In {\em ECCV}, 2018.

\bibitem{Bulò_2018_CVPR}
S.~Rota~Bulò, L.~Porzi, and P.~Kontschieder.
\newblock In-place activated batchnorm for memory-optimized training of dnns.
\newblock In {\em CVPR}, 2018.

\bibitem{Sajjadi_2018_CVPR}
M.~S.~M. Sajjadi, R.~Vemulapalli, and M.~Brown.
\newblock Frame-recurrent video super-resolution.
\newblock In {\em CVPR}, 2018.

\bibitem{shen2018factorized}
Z.~Shen, M.~Zhang, S.~Yi, J.~Yan, and H.~Zhao.
\newblock Factorized attention: Self-attention with linear complexities.
\newblock {\em arXiv:1812.01243}, 2018.

\bibitem{Shi_2016_CVPR}
W.~Shi, J.~Caballero, F.~Huszar, J.~Totz, A.~P. Aitken, R.~Bishop, D.~Rueckert,
  and Z.~Wang.
\newblock Real-time single image and video super-resolution using an efficient
  sub-pixel convolutional neural network.
\newblock In {\em CVPR}, 2016.

\bibitem{shuai2017dag}
B.~Shuai, Z.~Zuo, B.~Wang, and G.~Wang.
\newblock Scene segmentation with dag-recurrent neural networks.
\newblock {\em PAMI}, 2017.

\bibitem{igcv3}
K.~Sun, M.~Li, D.~Liu, and J.~Wang.
\newblock Igcv3: Interleaved low-rank group convolutions for efficient deep
  neural networks.
\newblock {\em arXiv:1806.00178}, 2018.

\bibitem{vaswani2017attention}
A.~Vaswani, N.~Shazeer, N.~Parmar, J.~Uszkoreit, L.~Jones, A.~N. Gomez,
  {\L}.~Kaiser, and I.~Polosukhin.
\newblock Attention is all you need.
\newblock In {\em NIPS}, 2017.

\bibitem{WahCUB_200_2011}
C.~Wah, S.~Branson, P.~Welinder, P.~Perona, and S.~Belongie.
\newblock {The Caltech-UCSD Birds-200-2011 Dataset}.
\newblock Technical report, 2011.

\bibitem{wang2017understanding}
P.~Wang, P.~Chen, Y.~Yuan, D.~Liu, Z.~Huang, X.~Hou, and G.~Cottrell.
\newblock Understanding convolution for semantic segmentation.
\newblock In {\em WACV}, 2018.

\bibitem{wang2018non}
X.~Wang, R.~Girshick, A.~Gupta, and K.~He.
\newblock Non-local neural networks.
\newblock In {\em CVPR}, 2018.

\bibitem{wu2018treestructured}
T.~Wu, S.~Tang, R.~Zhang, J.~Cao, and J.~Li.
\newblock Tree-structured kronecker convolutional network for semantic
  segmentation.
\newblock {\em arXiv:1812.04945}, 2018.

\bibitem{xiao2018unified}
T.~Xiao, Y.~Liu, B.~Zhou, Y.~Jiang, and J.~Sun.
\newblock Unified perceptual parsing for scene understanding.
\newblock In {\em ECCV}, 2018.

\bibitem{igcv2}
G.~Xie, J.~Wang, T.~Zhang, J.~Lai, R.~Hong, and G.-J. Qi.
\newblock Igcv $2 $: Interleaved structured sparse convolutional neural
  networks.
\newblock {\em arXiv:1804.06202}, 2018.

\bibitem{Yang_2018_CVPR}
M.~Yang, K.~Yu, C.~Zhang, Z.~Li, and K.~Yang.
\newblock Denseaspp for semantic segmentation in street scenes.
\newblock In {\em CVPR}, 2018.

\bibitem{yang2019deeperlab}
T.-J. Yang, M.~D. Collins, Y.~Zhu, J.-J. Hwang, T.~Liu, X.~Zhang, V.~Sze,
  G.~Papandreou, and L.-C. Chen.
\newblock Deeperlab: Single-shot image parser.
\newblock {\em arXiv:1902.05093}, 2019.

\bibitem{yu2018bisenet}
C.~Yu, J.~Wang, C.~Peng, C.~Gao, G.~Yu, and N.~Sang.
\newblock Bisenet: Bilateral segmentation network for real-time semantic
  segmentation.
\newblock In {\em ECCV}, 2018.

\bibitem{Yu_2018_CVPR}
C.~Yu, J.~Wang, C.~Peng, C.~Gao, G.~Yu, and N.~Sang.
\newblock Learning a discriminative feature network for semantic segmentation.
\newblock In {\em CVPR}, 2018.

\bibitem{yue2018cgnl}
K.~Yue, M.~Sun, Y.~Yuan, F.~Zhou, E.~Ding, and F.~Xu.
\newblock Compact generalized non-local network.
\newblock In {\em NIPS}, 2018.

\bibitem{OCNet}
Y.~Yuhui and J.~Wang.
\newblock Ocnet: Object context network for scene parsing.
\newblock {\em arXiv:1809.00916}, 2018.

\bibitem{clcnet}
D.-Q. Zhang.
\newblock clcnet: Improving the efficiency of convolutional neural network
  using channel local convolutions.
\newblock In {\em CVPR}, 2018.

\bibitem{Zhang_2018_CVPR}
H.~Zhang, K.~Dana, J.~Shi, Z.~Zhang, X.~Wang, A.~Tyagi, and A.~Agrawal.
\newblock Context encoding for semantic segmentation.
\newblock In {\em CVPR}, 2018.

\bibitem{zhang2019co}
H.~Zhang, H.~Zhang, C.~Wang, and J.~Xie.
\newblock Co-occurrent features in semantic segmentation.
\newblock In {\em CVPR}, 2019.

\bibitem{Zhang_2017_ICCV}
R.~Zhang, S.~Tang, Y.~Zhang, J.~Li, and S.~Yan.
\newblock Scale-adaptive convolutions for scene parsing.
\newblock In {\em ICCV}, 2017.

\bibitem{igcv1}
T.~Zhang, G.-J. Qi, B.~Xiao, and J.~Wang.
\newblock Interleaved group convolutions.
\newblock In {\em ICCV}, 2017.

\bibitem{Zhang_2018}
X.~Zhang, X.~Zhou, M.~Lin, and J.~Sun.
\newblock Shufflenet: An extremely efficient convolutional neural network for
  mobile devices.
\newblock {\em CVPR}, 2018.

\bibitem{zhao2017pyramid}
H.~Zhao, J.~Shi, X.~Qi, X.~Wang, and J.~Jia.
\newblock Pyramid scene parsing network.
\newblock In {\em CVPR}, 2017.

\bibitem{psanet}
H.~Zhao, Z.~Yi, L.~Shu, S.~Jianping, C.~C. Loy, L.~Dahua, and J.~Jia.
\newblock Psanet: Point-wise spatial attention network for scene parsing.
\newblock In {\em ECCV}, 2018.

\bibitem{zhou2017scene}
B.~Zhou, H.~Zhao, X.~Puig, S.~Fidler, A.~Barriuso, and A.~Torralba.
\newblock Scene parsing through ade20k dataset.
\newblock In {\em CVPR}, 2017.

\bibitem{zhou2018unet++}
Z.~Zhou, M.~M.~R. Siddiquee, N.~Tajbakhsh, and J.~Liang.
\newblock Unet++: {A} nested u-net architecture for medical image segmentation.
\newblock In {\em {MICCAI}}, 2018.

\end{thebibliography}
}

\end{document}